# Upper, Middle and Lower Region Learning for Facial Action Unit Detection

Yao Xia*
School of Science, Edith Cowan University
Email: y.xia@ecu.edu.au

**Abstract**
Facial action units (AUs) detection is fundamental to facial expression analysis. As AU occurs only in a small area of the face, region-based learning has been widely recognized useful for AU detection. Most region-based studies focus on a small region where the AU occurs. Focusing on a specific region helps eliminate the influence of identity, but bringing a risk for losing information. It is challenging to find balance. In this study, I propose a simple strategy. I divide the face into three broad regions, upper, middle, and lower region, and group AUs based on where it occurs. I propose a new end-to-end deep learning framework named three regions based attention network (TRA-Net). After extracting the global feature, TRA-Net uses a hard attention module to extract three feature maps, each of which contains only a specific region. Each region-specific feature map is fed to an independent branch. For each branch, three continuous soft attention modules are used to extract higher-level features for final AU detection. In the DISFA dataset, this model achieves the highest F1 scores for the detection of AU1, AU2, and AU4, and produces the highest accuracy in comparison with the state-of-the-art methods.

## 1 Introduction

Facial expression is defined as motions of the muscles beneath the facial skin [1]. It is a fundamental part of human emotion, and playing a critical role in human-machine interaction. Therefore, facial expression analysis has attracted more and more interest in the field of computer vision. Facial action coding system (FACS) [2] is one of the most popular facial expression coding systems and has been used in the analysis of emotion [3], depression [4], and measuring pain [5] etc. In FACS, the expression of the whole face is divided into different sub-expressions coded by action units (AUs), each of which is linked to the motion of a specific part of facial muscles. Hence, to detect AU should concern more about local information rather than global information.

Automatic AU detection is a popular topic in facial expression analysis. Many recent works, especially the deep learning-based methods, incorporated the global feature of the whole face as at least a part of the model. Nevertheless, the global feature will bring noise from various identities. There have been many studies that tried to eliminate the effect of identity [6] [7] [8], but the extra processing modules could bring additional error and computation cost.

The attention mechanism is a simple but effective way to make the model focus on local information. We expect machines can pay more attention to the most important part and ignore the less important part when seeing an image, just like the human visual system. It has been widely recognized that different regions in a face contribute differently to the model [9] [10] [11]. There have been many models adopting visual attention to enhance the performance of AU detection. However, some of them are heavy, and some need multiple pre-defined key points- based attention maps, thus increase the computation cost and the difficulty of training significantly. Furthermore, most region-based models focus on a small part of the whole image, so information on a larger scale was ignored. Besides, the channel attention, whose importance has been recognized for image classification, has been ignored by most of the studies of AU detection.

In this work, I propose a novel end-to-end deep learning framework named three regions based attention network (TRA- Net) for detecting AU. This model uses a SENet-50 [12] model pre-trained on the VGG Face2 dataset [13] as the backbone, and a decoder with skip connection is connected to the last feature map before the final average pooling layer to up-sample the feature map. Then, based on the central point of the face, three hard masks focusing on the upper, middle, and lower parts of the face are generated and multiplied with the feature map to produce three branches containing only upper, middle, and lower information of the face respectively. Three continuous blocks of CBAM are then used to extract local information further. Finally, for each branch, two fully connected layers are adopted to conduct AU prediction. Based on the characteristic of AU, each branch is designed for different groups of AU. For example, the upper branch is only used to detect AU1, AU2, and AU4. In addition, the TRA-Net does not use the residual connection in the attention block. By doing so, global information can be eliminated.

This work makes three main contributions. First, I propose an easy-to-train end-to-end deep learning framework named three regions based attention network (TRA-Net) for AU detection. On the benchmark dataset, DISFA, it outperforms the state-of-the-art methods in the F1 score of AU1, AU2 (F1 score of AU4 is very close to the state-of-the-art). TRA-Net also outperforms the state-of-the-art model in accuracy. Second, I use both hard mask and soft attention mask to extract critical features without residual connection, which means the global information is discarded as much as possible. My experiments demonstrate that the model without the residual connection outperforms model with that, implying that using only local information is helpful for AU detection. Third, I use Squeeze-and-Excitation (SE) modules to learn channel attention in all modules of TRA-Net, and demonstrate that using channel attention is the same important as using spatial attention.

## 2 Related works

It has been widely recognized that AU is sub-region-specific. Many studies have adopted the idea of region-based learning for AU detection (or emotion analysis).

For example, Zhao et al. [10] proposed a joint patch and multi-label learning (JPML). JPML defined a sparse subset of facial patches based on crucial landmarks and used patches learning to detect AU. Zhong et al. [14] proposed a two-stage multi-task, sparse learning (MTSL). In this model, the whole face image was divided into equal-size patches in different scales, and then the common patches and the specific patches across expressions were learned for the emotion classification. Fernandez et al. [15] proposed an attention-based network for emotion recognition, which multiply the feature map with a soft mask generated by an encoder-decoder with a skip connection structure. Zhao et al. [11] proposed Deep Region and Multi-label Learning (DRML) for AU detection. DRML used the convolution layer with unshared weights for learning, i.e., after dividing a feature into patches with equal size, the kernel weights only shared within the same local patches. Thus different patches were learned independently. Li et al. [16] proposed enhancing and cropping net (EAC-net), which used VGG net as the backbone and incorporated an enhancing module and a cropping module. The enhancing module adopted a pre-defined mask generated from crucial landmarks associated with specific AUs to enhance the representation of essential regions according to AUs, and then the cropping module was used to learn the feature from each AU specific area. Shao et al. [17] developed a framework called JAA-Net to do face alignment and AU detection in the same end-to-end network. JAA-Net used convolution layers with unshared weights to extract global information and then trained the face alignment module. After that, the predicted landmarks were used to generate pre-defined attention maps for each AU, then after a refine step, the refined attention maps were multiplied by the corresponding feature maps, and the final output was used to predict the occurrence of AU. Corneanu et al. [18] proposed a Deep Structure Inference Network (DSIN) that can learn the relationships of AU classes by information passing algorithms between AU predictions. Shao et al. [19] proposed an Attention and Relation Learning (ARL) network. ARL adopted both spatial and channel attention and also used a pixel-level relation learning for AU detection and intensity estimation. Li et al. [20] proposed semantic relationship embedded representation learning (SRERL) that can learn AU semantic to enhance the features representation of facial regions. Niu et al. [21] proposed Local Relationship Learning with Person-specific Shape Regularization (LP-Net) for AU detection, LP-Net could use local information and the relationship of individual local face regions to improve the AU detection robustness, and used person-specific shape regularization to reduce the influence of the diverse baseline AU intensity. Shao et al. [22] proposed Spatial-Temporal Relation and Attention Learning for Facial Action Unit Detection (STRAL). STRAL used a framework like ARL to extract information for each frame and then uses a spatial-temporal relation learning module for learning the AU relations.

## 3 TRA-Net for AU detection

The structure of TRA-Net is shown in Figure 1. The model uses a pre-trained SENet50 [12] as backbone for extracting global information. SENet 50 is a classical Resnet-50 network containing Squeeze-and-Excitation (SE) module. In the convolution layer, each channel contributes differently to the model, but many classical network frameworks do not take channel-wise relationship into account. SE is a lightweight gating mechanism to learn channel attention and can be easily added into convolution layer. SE can build on a transfomation from input to a feature map. SE consists of two steps. The first is squeezing, which means squeezing global spatial information into a channel descriptor. For example, for feature $U \in R^{C \times H \times W}$ the spatial dimensions $H \times W$ were squeezed in to 1 using a global average pooling function as following:

$$z_c = \frac{1}{H \times W} \sum_{i=1}^{H} \sum_{j=1}^{W} u_c(i,j) \quad (1)$$

where $u_c$ denotes the $c$-th element of feature $U$, $z_c$ denotes the $c$-th element of feature $Z \in R^{C \times 1 \times 1}$.

Then $Z$ is fed into two dimensionality-reduction fully connected (FC) layers with reduction ratio $r$,

$$s = \sigma(W_2 \delta(W_1 z)) \quad (2)$$

where $W \in R^{C/r \times C}$ represents the FC layer (1 & 2), $z$ is element of $Z$, $\delta$ denotes a Relu function and $\sigma$ is a sigmoid function.

The output $S \in R^{C \times 1 \times 1}$ is expanded into $S_e \in R^{C \times H \times W}$, where the value of every pixel in the same channel is equal. $S_e$ can be regarded as the channel-wise weights, and the output $O \in R^{C \times H \times W}$ of SE module with residual structure can be calculated as follows:

$$O = (S_e \otimes U) \oplus U \quad (3)$$

where $\otimes$ denotes the element-wise multiply, $\oplus$ denotes the element-wise addition.

Besides the first convolution layer with kernel size 7, SE is used in each block of ResNet-50 to form the SENet-50 framework for TAR-Net. Then the weights of SENet-50 pre-trained on VGGface-2 dataset [13] were loaded.

The backbone is used to extract global information. After that, the last layer before the global average pooling layer is up-sampled by a decoder having the same shape with the corresponding part in the last three blocks, where the skip connection is used to keep the lower-level features. The SE module is also added to the decoder to learn channel attention.

After upsampling, a hard attention module consisting of three hard masks is adopted to divide the feature map into three parts, upper region, middle region and lower region. In the frame pre-processing step, the landmark located in the middle of nose tip and nose root is used as the centre point to perform a similar transformation, so that in every processed frame, the middle point is always close to that landmark. Let $I \in R^{C \times H \times W}$ denotes the input feature map with centre point $C$ ($x$, $y$) and the hard masks are $M_{h\_up}$, $M_{h\_mid}$, $M_{h\_low}$. The hard masks have the same dimensions $H$ and $W$ with the input feature and contain only integrates $0$ and $1$. The hard masks split the feature map horizontally, the bottom (denotes the bottom of the region with value one in a hard mask) of $M_{h\_up}$ and the top (denotes the top of the region with value one in a hard mask) of $M_{h\_low}$ are corresponding to $y$, and the top and the bottom of $M_{h\_mid}$ are $y + H/4$ and $y - H/4$ respectively.

$$I_{hm} = M_h \otimes I \quad (4)$$

where $I_{hm}$ is the masked feature map, $M_h$ is an expanded hard mask. The decoded feature map is fed into the hard mask module to generate three masked features, each of which is then fed to a specific branch.

Let $R_{up}$, $R_{mid}$ and $R_{low}$ represent the masked feature maps and regarded as inputs of upper region branch, middle region branch and lower region branch respectively. $R_{up}$, $R_{mid}$ and $R_{low}$ are fed into three continuous soft mask attention blocks in order to gradually refine the attention and learn higher-level features. Here the convolution block attention module (CBAM) [24] is used. CBAM adopts SE blocks for learning both spatial and channel attention. To learn the spatial attention, the idea is similar. Firstly the channel dimension $C$ is squeezed into 1 while dimensions $H$ and $W$ keep unchanged. After squeezing, a convolution block is used to learn the spatial feature, and then the learned feature map is expanded back to $C$ with same values.

Finally element-wise multiply the expanded spatial mask with the input feature map to get the masked feature map. In addition, CBAM uses both max pooling and average pooling to perform squeezing, and add the output of both squeezed parts at the final step. Two convolution layers with residual connection and a max pooling layer are connected to the end of first two CBAM blocks, and two convolution layers and a average pooling layer are connected to the last CBAM block. Finally, the output of the average pooling layer is fed to two FC layers to predict AU. For each branch, multi-label should be predicted. Upper region branch is responsible for detecting AU1, AU2, AU4, middle region branch is for AU6, AU9 and lower region branch is for AU12, AU25, AU 26.

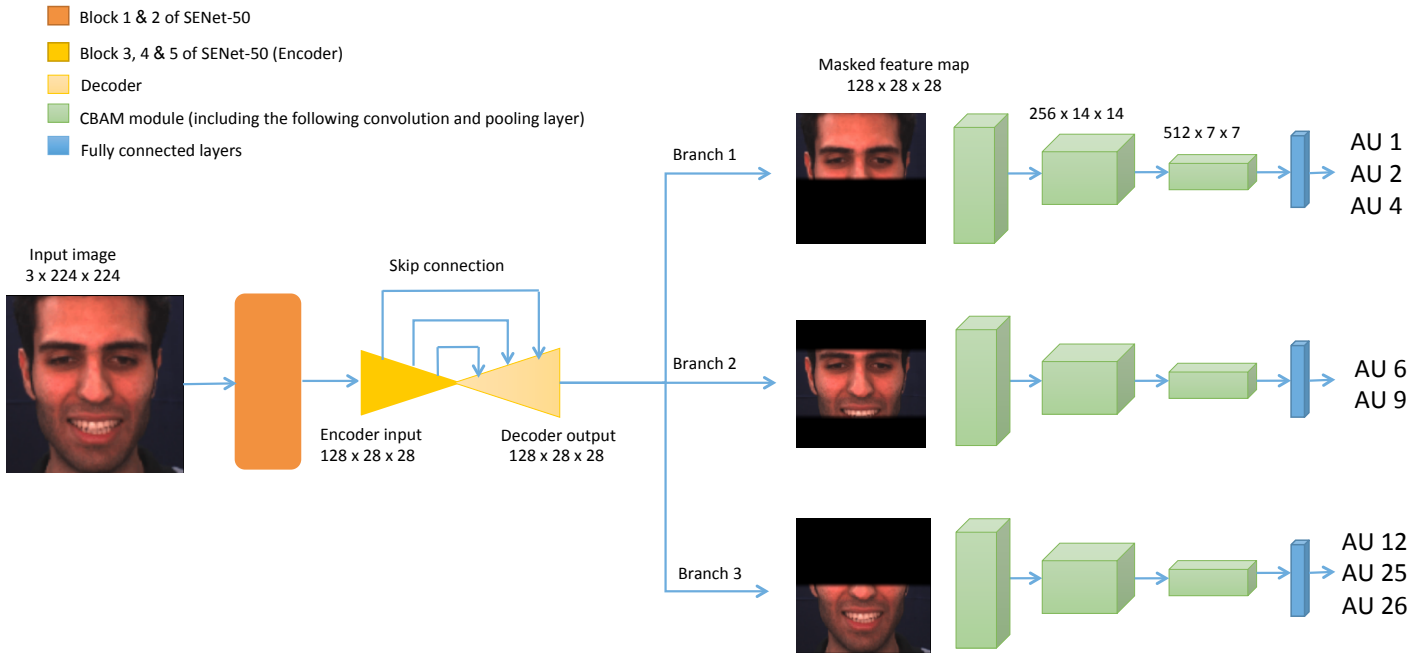

**Figure 1.** The architecture of TRA-Net. The blue arrow denotes the direction of forwarding propagation. The backbone is split into two parts. One part is Block 1 and 2, which are frozen during training. The other part is Block 3, 4 and 5. These three modules are reagarded as an encoder and than connected to a decoder. The decoder is used for up-sampling, so it has only one convolutional layer at each corresponding block. The skip connection is used to keep the low-level information. Two convolution layers and a max pooling layer is connected to the ends of first two CBAM modules, and two convolution layers and a average pooling layer are connected to the end of the first CBAM module.

## 4 Experiments
### 4.1 Dataset and settings:
TAR-Net was evaluated on DISFA dataset [25]. DISFA contains videos from 27 individuals. AUs were coded frame-by-frame, and 66 landmarks were annotated by AAM (bad annotations has been marked). Following the settings of [11], I used a 3-fold cross-validation scheme to evaluate the performance of TAR-Net, and the frame with AU intensity equal or higher than C-level was regarded as positive, and others are treated as negative.

**Implementation details:** For each frame, a similarity transformation was performed for face alignment. Two rules were used: (1) make the landmark located in the middle of the nose tip and nose root close to the centre of the frame; (2) make the landmark located in the middle of the lower jaw close the bottom of the frame. By doing so, the shape of the face can keep unchanged, and the face was put straight. In addition, this also facilitates generating hard masks. The output shape of similarity transformation was 3 X 224 X 224. In order to enhance diversity, a random horizontally flip strategy was used during training. The labels in

DISFA is highly imbalanced, which effects the training significantly. Thus I adopted a random sampling strategy to generate the training batches. The dataset was split into training and test set based on identity. In a 3-fold setting, frames from 18 individuals were used as the training set, and frames from the other 9 individuals were treated as the testing set. Firstly the weight for each positive AU label was calculated for the training set, then the probability for choosing a frame with a positive label for a specific AU was calculated based on its weight. For each batch, 64 frames were randomly selected from the training set, and 1 epoch contained 100 batches. TRA-Net was trained using Pytorch framework. Stochastic gradient descent (SGD), with momentum of 0.9, weight decay of 0.001, initial learning rate of 0.001, was used to optimize the network. As the backbone SENet-50 has been pre-trained on VGG face 2 datasets, I froze the first two blocks of the backbone and trained the rest part of TRA-Net. Multilabel binary cross-entropy was used as the loss function, and losses from the upper branch, middle branch, and lower branch were summed as the final loss for backpropagation. I started to evaluate the performance on the test set at the end of every epoch when the model was close to convergence, and stop training when the performance did not increase after 10 epochs.

**Evaluation metrics:** Following most related works, F1 socre and accuracy were used for evaluation.

### 4.2 Comparison with atate-of-the-art methods

I compared TRA-Net against other AU detection methods, under the same 3-fold cross validation setting, including CMS [24], LP [21], DRML [11], EAC [16], DSIN [18], JAA [17], SRERL [20], ARL [19], STRAL [22]. The restuls were in Table 1.

Table 1. F1-frame and accuracy for 8 Aus on DISFA

| AU | F1-Frame | | | | | | | | | | Accuracy | | | | | |
|---|---|---|---|---|---|---|---|---|---|---|---|---|---|---|---|---|
| | CMS | LP | DRML | EAC | DSIN | JAA | SRERL | ARL | STRAL | **TRA** | CMS | EAC | JAA | ARL | STRAL | **TRA** |
| 1 | 40.2 | 29.9 | 17.3 | 41.5 | 42.4 | 43.7 | 45.7 | 43.9 | 52.2 | **56.9** | 91.6 | 85.6 | 93.4 | 92.1 | 94.9 | **97.3** |
| 2 | 44.3 | 24.7 | 17.7 | 26.4 | 39.0 | 46.2 | 47.8 | 42.1 | 47.4 | **61.5** | 94.7 | 84.9 | 96.1 | 92.7 | 93.5 | **97.9** |
| 4 | 53.2 | **72.7** | 37.4 | 66.4 | 68.4 | 56.0 | 59.6 | 63.6 | 68.9 | 72.6 | 79.9 | 79.1 | 86.9 | 88.5 | 89.5 | **93.5** |
| 6 | **57.1** | 46.8 | 29.0 | 50.7 | 28.6 | 41.4 | 47.1 | 41.8 | 47.8 | 45.2 | 82.6 | 69.1 | 91.4 | 91.6 | 90.4 | **96.2** |
| 9 | 50.3 | 49.6 | 10.7 | **80.5** | 46.8 | 44.7 | 45.6 | 40.0 | 56.7 | 46.0 | 95.2 | 88.1 | 95.8 | 95.9 | 96.8 | **97.6** |
| 12 | 73.5 | 72.9 | 37.7 | **89.3** | 70.8 | 69.6 | 73.5 | 76.2 | 72.5 | 67.5 | 87.8 | 90.0 | 91.2 | 93.9 | 92.4 | **94.2** |
| 25 | 81.1 | 93.8 | 38.5 | 88.9 | 90.4 | 88.3 | 84.3 | **95.2** | 91.3 | 89.5 | 86.3 | 80.5 | 93.4 | 97.3 | 94.9 | **95.9** |
| 26 | 59.7 | 65.0 | 20.1 | 15.6 | 42.2 | 58.4 | 43.6 | 66.8 | **67.6** | 33.1 | 80.7 | 64.8 | 93.2 | **94.3** | 94.0 | **94.3** |
| **Avg** | 57.4 | 56.9 | 26.7 | 48.5 | 53.6 | 56.0 | 55.9 | 58.7 | **63.0** | 59.0 | 87.3 | 80.6 | 92.7 | 93.3 | 93.3 | **95.9** |

For DISFA dataset, TRA-Net improves 9% for F1 score of AU1 and 28.7% for F1 score of AU2 respectively over the state-of-the-art method. The F1 score of AU4 is very close to the state-of-the-art model. For the rest of AUs, TRA-Net does not bring significant improvement. Detection of AU1, AU2, AU4 is outputted by the upper region branch, which implies that TRA-Net brings significant improvement for detecting AUs occurring on the upper face. TRA-Net brings overall improvement in accuracy. It should be noted that DISFA is a highly imbalanced dataset, and most labels are negative, so the high accuracy mainly results from correct prediction for the negative samples. It should also be noted that TRA-Net is trained on DISFA dataset directly, which demonstrates its ability to process real-world data.

### 4.3 Ablation studies

In order to evaluate the effectiveness of each module of TRA-Net, I also conducted experiments on the model with one or more modules removed. The results on F1 score were shown in Table 2. SENet-50 denotes the backbone model, HM denotes the hard mask module, CBAM denotes the three continuous CBAM modules, CBAM w/o AT represents CBAM module without channel attention block and CBAM w/o ST represents CBAM module without spatial attention block.

Table 2. F1-frame for different variants of TRA-Net

| Methods | AU 1 | AU 2 | AU 4 | AU 6 | AU 9 | AU 12 | AU 25 | AU 26 | Avg |
|---|---|---|---|---|---|---|---|---|---|
| SENet-50 | 27.4 | 25.8 | 70.3 | 29.3 | 36.6 | 64.1 | 88.7 | 34.0 | 47.0 |
| SENet-50 + CBAM | 35.4 | 39.4 | 66.9 | 31.2 | 39.4 | 67.4 | 89.9 | 30.0 | 49.9 |
| SENet-50 + HM | 33.0 | 37.0 | 58.1 | 46.7 | 50.0 | 68.7 | 90.2 | 31.5 | 51.9 |
| SENet-50 + HM + CBAM w/o AT | 45.2 | 47.8 | 66.5 | 41.1 | 41.3 | 72.3 | 87.1 | 41.0 | 55.3 |
| SENet-50 + HM + CBAM w/o ST | 45.2 | 45.8 | 65.8 | 47.3 | 55.1 | 70.5 | 89.4 | 33.3 | 56.2 |
| TRA-Net with residual structure | 48.9 | 49.4 | 63.1 | 46.8 | 47.3 | 70.2 | 89.8 | 34.8 | 56.3 |

**SENet-50:** From the results, I found the performance of SENet-50 for AU 12, 25, 26 is very close to TRA-Net. This implied that the attention modules did not bring any improvement to the detection of AUs on the lower face.

**Hard mask:** Hard mask is a crucial component of TRA-Net. It divided the image into different regions, eliminating unnecessary parts for detecting specific AUs. After removing the hard mask module, the model performance dropped significantly.

**CBAM:** CBAM module was used to extract further information on the feature map masked by hard attention. It applied both spatial and channel attention to the model. From the results, I found it is also a necessary part of TRA-Net. I also removed the channel and spatial attention modules in CBAM respectively, and found the performance has dropped without any of them. Therefore the channel attention is as same important as spatial attention in CBAM.

**Residual connection:** I tried to add a residual connection in CBAM modules to keep the information of the input feature map. I found that the residual only helped the model convergent more quickly, but made the performance worse.

## 5 Conclusion

In this work, I propose a new end-to-end deep learning framework named three regions based attention network (TRA-Net) for AU detection. TRA-Net uses a pre-trained SENet-50 as the backbone for extracting face global features, and then a hard mask module to divide the face into three region branches (upper, middle, lower). After the hard attention, three independent branches are generated, and the corresponding inputs are fed into three continuous soft attention modules (CBAM) for final AU prediction. My model outperforms the state-of-the-art methods on F1 score for AUs occurring in the upper face and outperforms the state-of-the-art on accuracy for detection of all AUs. I used a randomly sampling strategy to deal with the label imbalance problem and achieved good performance.